\pdfoutput=1
\documentclass[11pt]{article}
\usepackage{acl}

\usepackage{times}

\usepackage[utf8]{inputenc} % allow utf-8 input
\usepackage[T1]{fontenc}    % use 8-bit T1 fonts
\usepackage{hyperref}       % hyperlinks
\usepackage{url}            % simple URL typesetting
\usepackage{booktabs}       % professional-quality tables
\usepackage{amsfonts}       % blackboard math symbols
\usepackage{nicefrac}       % compact symbols for 1/2, etc.
\usepackage{microtype}      % microtypography
\usepackage{xcolor}         % colors
\usepackage{standalone}
\usepackage{latexsym}
\usepackage{amsmath}
\usepackage{amssymb}
\usepackage{amsthm}
\usepackage{graphicx}
\usepackage{subcaption}
\usepackage{array}
\usepackage{tabu}
\usepackage{makecell}
\usepackage{paralist}
\usepackage{cases}
\usepackage{diagbox}
\usepackage{enumitem}
\usepackage{soul}
\usepackage{multirow}
\usepackage{verbatim}
\usepackage{tabulary}
\usepackage{booktabs}
\usepackage[mathscr]{euscript}
\usepackage{mathtools}
\usepackage{algorithm}
\usepackage{algpseudocode}
\usepackage{stmaryrd}
\usepackage{tikz-dependency}
\usetikzlibrary{automata,decorations.markings,arrows,positioning,matrix,calc,patterns,angles,quotes,calc}
\usepackage{adjustbox}
\usepackage{tabularx}
\usepackage{xspace}
\usepackage{tabulary}
\usepackage{afterpage}
\usepackage{hyperref}
\usepackage{url}
\usepackage{bm}
\usepackage{color}
\usepackage{graphicx}
\usepackage{slashbox}
\usepackage[toc,page]{appendix}
\usepackage{makecell}
\usepackage{boldline}

\definecolor{orange}{rgb}{1,0.5,0}
\definecolor{mdgreen}{rgb}{0.05,0.6,0.05}
\definecolor{mdblue}{rgb}{0,0,0.7}
\definecolor{dkblue}{rgb}{0,0,0.5}
\definecolor{dkgray}{rgb}{0.3,0.3,0.3}
\definecolor{slate}{rgb}{0.25,0.25,0.4}
\definecolor{gray}{rgb}{0.5,0.5,0.5}
\definecolor{ltgray}{rgb}{0.7,0.7,0.7}
\definecolor{purple}{rgb}{0.7,0,1.0}
\definecolor{lavender}{rgb}{0.65,0.55,1.0}

\definecolor{mypurple}{RGB}{111,61,121}
\definecolor{myblue}{RGB}{46,88,180}
\definecolor{myred}{RGB}{181,68,106}
\definecolor{myyellow}{RGB}{204,143,55}

\newcommand{\ensuretext}[1]{#1}
\newcommand{\marker}[2]{\ensuremath{^{\textsc{#1}}_{\textsc{#2}}}}
\newcommand{\arkcomment}[3]{\ensuretext{\textcolor{#3}{[#1 #2]}}}
\renewcommand{\arkcomment}[3]{}  % uncomment for submission

\newcommand{\nikos}[1]{\arkcomment{\marker{N}{P}}{#1}{mdgreen}}

\newcommand{\hao}[1]{\arkcomment{\marker{H}{P}}{#1}{purple}}
\newcommand{\jungo}[1]{\arkcomment{\marker{J}{K}}{#1}{brown}}

\newcommand{\rev}[1]{{#1}}

\newcommand{\term}[1]{\textbf{#1}} % term being defined

\newcommand{\interalia}[1]{\citep[\emph{inter alia}]{#1}}

\newcommand{\relu}{\operatorname{ReLU}}

\DeclareSymbolFont{extraup}{U}{zavm}{m}{n}
\DeclareMathSymbol{\vardiamond}{\mathalpha}{extraup}{87}

\newcolumntype{L}[1]{>{\raggedright\let\newline\\\arraybackslash\hspace{0pt}}m{#1}}
\newcolumntype{C}[1]{>{\centering\let\newline\\\arraybackslash\hspace{0pt}}m{#1}}
\newcolumntype{R}[1]{>{\raggedleft\let\newline\\\arraybackslash\hspace{0pt}}m{#1}}

\newtheorem{theorem}{Theorem}

\theoremstyle{definition}

\newtheorem{example}[theorem]{Example}
\theoremstyle{remark}

\algrenewcommand{\algorithmiccomment}[1]{\leavevmode$\triangleright$ #1}

\setul{1pt}{.4pt}

\usepackage[shortcuts]{extdash}  % for non-breaking hyphens. Stack Overflow says it's important for this to be the last package loaded.

\usepackage{blindtext}
\usepackage{graphicx}
\usepackage{capt-of}
\usepackage{booktabs}
\usepackage{varwidth}
\newsavebox\tmpbox
\usepackage{amsmath,amsfonts,bm}

\def\eqref#1{equation~\ref{#1}}
\def\floor#1{\lfloor #1 \rfloor}
\def\1{\bm{1}}

\def\vzero{{\mathbf{0}}}

\def\ve{{\mathbf{e}}}

\def\vk{{\mathbf{k}}}

\def\vq{{\mathbf{q}}}

\def\vv{{\mathbf{v}}}
\def\vw{{\mathbf{w}}}
\def\vx{{\mathbf{x}}}
\def\vy{{\mathbf{y}}}
\def\vz{{\mathbf{z}}}
\def\vphi{{\boldsymbol{\phi}}}
\def\valpha{{\boldsymbol{\alpha}}}

\def\vexp{{\boldsymbol{\exp{}}}}

\def\mI{{\mathbf{I}}}

\def\mK{{\mathbf{K}}}

\def\mM{{\mathbf{M}}}

\def\mU{{\mathbf{U}}}
\def\mV{{\mathbf{V}}}
\def\mW{{\mathbf{W}}}
\def\mX{{\mathbf{X}}}

\DeclareMathAlphabet{\mathsfit}{\encodingdefault}{\sfdefault}{m}{sl}
\SetMathAlphabet{\mathsfit}{bold}{\encodingdefault}{\sfdefault}{bx}{n}

\def\gO{{\mathcal{O}}}

\def\sR{{\mathbb{R}}}

\newcommand{\softmax}{\operatorname{softmax}}

\newcommand{\attn}{\operatorname{attn}}

\newcommand{\model}{\textsc{Abc}\xspace}
\newcommand{\modelmlp}{\textsc{Abc}$_\text{MLP}$\xspace}

\newcommand{\modelrandom}{\textsc{Abc}$_{\text{RD}}$\xspace}

\newcommand{\resolved}[1]{}
\newcommand{\base}[0]{\textsc{Base}\xspace}

\newcommand{\com}[1]{}
\title{\model: Attention with Bounded-Memory Control}

\author{\textbf{Hao Peng}$^\spadesuit$ \quad
        \textbf{Jungo Kasai}$^\spadesuit$ \quad
        \textbf{Nikolaos Pappas}$^{\bigstar*}$ \quad
        \textbf{Dani Yogatama}$^\clubsuit$ \quad
        \textbf{Zhaofeng Wu}$^{\diamondsuit}$\thanks{\hspace{1mm}This work was done while Zhaofeng Wu and Nikolaos Pappas were at the University of Washington.}\\
        \textbf{Lingpeng Kong}$^{\vardiamond}$\quad
        \textbf{Roy Schwartz}$^\heartsuit$ \quad
        \textbf{Noah A. Smith}$^{\spadesuit\diamondsuit}$\\
  $^\spadesuit$Paul G. Allen School of Computer Science \& Engineering,
  University of Washington\\
  $^\bigstar$Amazon Web Services\quad $^\clubsuit$DeepMind\quad
  $^\diamondsuit$Allen Institute for Artificial Intelligence\\ 
  $^\heartsuit$School of Computer Science \& Engineering, Hebrew University of Jerusalem\\
  $^\vardiamond$Department of Computer Science, The University of Hong Kong \\ 
  {\tt \{hapeng,jkasai,npappas,zfw7,nasmith\}@cs.washington.edu} \\
  {\tt dyogatama@deepmind.com, lpk@cs.hku.hk}\\
  {\tt roy.schwartz1@mail.huji.ac.il}
}
\begin{document}

\maketitle

\begin{abstract}

Transformer architectures have achieved state-of-the-art results on a variety of natural language processing (NLP) tasks.
However, their attention mechanism comes with a quadratic complexity in sequence lengths,
making the computational overhead prohibitive, especially for long sequences.
Attention context can be seen as a random-access memory with each token taking a slot.
Under this perspective, the memory size grows linearly with the sequence length, and so does the overhead of reading from it.
One way to improve the efficiency is to bound the memory size.
We show that disparate approaches\resolved{\nikos{"for bounding memory size" or "for this purpose"}\hao{i think it's fine as is}} can be subsumed into one abstraction, 
\textbf{a}ttention with \textbf{b}ounded-memory \textbf{c}ontrol (\model),
and they vary in their  \emph{organization} of the memory.
\model reveals  new, unexplored possibilities.
First, it connects several efficient attention variants that would otherwise seem distinct.
Second, this abstraction gives new insights---an established approach~\citep{wang2020linformer} previously thought to \textit{not} be applicable in causal attention, \emph{actually is}.
Last, we present a new instance of \model, which draws inspiration from existing \model approaches,
but replaces their heuristic memory-organizing functions
with a \emph{learned}, contextualized one.
Our experiments on language modeling, machine translation, and masked language model finetuning
show that our approach outperforms previous efficient attention models;
compared to strong transformer baselines, it significantly improves the inference time and space
efficiency with no or negligible accuracy loss.
\end{abstract}
\section{Introduction}

Transformer architectures are now central in 
natural language processing~\citep{vaswani2017attention}.
They rely on the attention mechanism~\citep{bahdanau2015attention} to contextualize the input.
The context can be seen as a random access \term{memory} whose size linearly grows with the sequence length;
each query reads from it using a softmax-normalized linear combination, with overhead linear in the memory size.
This amounts to a quadratic complexity overall, making transformers' computational overhead prohibitive, especially for long sequences. 

One way to improve attention's efficiency is to bound its memory size.
Imposing a constant-sized constraint over the memory ensures that reading from it has constant time and space overhead,
yielding a linear overall complexity in sequence lengths. 
This is in fact a common strategy adopted by several recent works. In this work, we show that some of these works 
are closely connected in ways that, to date, have gone unremarked.
We propose \textbf{a}ttention with \textbf{b}ounded-memory \textbf{c}ontrol (\model),
a unified abstraction over them.
In \model, constant-sized memories are organized with various control strategies, e.g.,
induced from heuristic patterns~\interalia{beltagy2020longformer,zaheer2020big,ainslie2020etc,Rae2020Compressive}, 
locality assumptions~\citep{parmar2018image,liu2018generating}, 
or positions~\citep{wang2020linformer}.

These strategies, by and large, are ``context-agnostic.'' 
In response to this, we propose \modelmlp, a particular instance of \model that
learns a contextualized control strategy from data.
Specifically, \modelmlp uses a neural network to determine how to store each token into the memory (if at all).
Compared to previous bounded-memory models, it strikes a better trade-off between accuracy and efficiency:
controlling for the accuracy, \modelmlp can get away with much smaller memory sizes.

\model models (including \modelmlp) come with a linear complexity
in sequence lengths,
and admit recurrent computation graphs in causal attention (self-attention over the prefix).
Therefore they are appealing choices in a variety of applications, including text encoding, language modeling and text generation.
This leads to a surprising finding.
Linformer~\citep{wang2020linformer}, an established  efficient attention method,
was previously thought \emph{not} to be applicable in causal attention or autoregressive decoding~\citep{tay2020}.
Through the \model view, we show that it actually \emph{is},
\rev{and achieves competitive performance in our machine translation experiments.}

\model connects existing models that would otherwise seem distinct,
reveals new insights into established methods,
and inspires new efficient attention architectures.
We explore its applications in transformers, as a drop-in substitute for the canonical softmax attention.
\model offers a novel lens that can help
future research in the analysis of transformers,
where the theoretical insights are still catching up with  empirical success.
Experiments on language modeling, machine translation, and masked language model finetuning 
show that our \modelmlp model outperforms previous \model approaches in accuracy
with a much smaller memory size.
Compared to the strong transformer baseline,
\modelmlp achieves a significant speedup and memory savings at inference time,
with no or negligible accuracy loss.
The efficiency improvements are more prominent for long sequences,
suggesting that the asymptotic savings are even more appealing in applications
involving long sequences.
We release our code at \url{https://github.com/Noahs-ARK/ABC}.

\section{An Outer-Product View of Attention}\label{sec:background}

This section presents our outer-product memory perspective of attention,
which allows for a smooth transition to later discussion.

In attention,
a sequence of \term{queries} $\{\vq_i\}_{i=1}^{N}$
attend to a \term{memory} with $N$ slots,
each storing a \term{key} and \term{value} pair:
$\mK=[\vk_1, \dots, \vk_N]^\top, \mV=[\vv_1, \dots, \vv_N]^\top\in\sR^{N\times d}$.\footnote{
The number of queries and key-value pairs may differ, e.g., in
the cross attention of a sequence-to-sequence model.
}
Query $\vq$ reads from the memory using a softmax-normalized linear combination,
producing a $d$-dimensional vector: 
\begin{align}
\attn(\vq, \{\vk_i\}, \{\vv_i\})
    = \mV^\top\softmax\left(\mK\vq\right).
\end{align}
This takes $\gO(N)$ time and space.
When the attention with $N$ queries can be parallelized (e.g., in text encoding),
it takes linear time and quadratic space; 
when it \emph{cannot} be (e.g., in decoding), 
it takes quadratic time and linear space. 

The memory can be equivalently represented as sums of vector outer products:
%\begin{align}\label{eq:attn_outer}
    $\mK = \mI \mK =  \sum_{i=1}^{N} \ve_i \otimes \vk_i$, $\mV =  \sum_{i=1}^{N} \ve_i \otimes \vv_i$.
%\end{align}
$\mI$ is the identity matrix, and $\otimes$ denotes the outer product: $[\vx\otimes\vy]_{i,j}=x_i y_j$. 
$N$-dimensional vectors $\{\ve_i\}$ form the standard basis: $\ve_i$
has the $i$th element being one and others zeros.
We can view $\ve_i$ as \term{control vectors} that determine where to store $\vk_i$ and $\vv_i$: 
\begin{align}
\begin{split}
\ve_i \otimes \vk_i
&=\big[\underbrace{0,\dots0}_{i-1}, 1, \underbrace{0, \dots, 0}_{N-i}\big]^\top\otimes \vk_i\\
&=\big[\underbrace{\vzero}_{d\times (i-1)}; \vk_i;\underbrace{\vzero}_{d\times(N-i)}\big]^\top. %\in \sR^{N\times d}.
\end{split}
\end{align}
The $N$-by-$d$ matrix on the last line has its 
$i$th row being $\vk_i^\top$ and all others zeros;
in this sense, $\vk_i$ is stored in the $i$th slot by $\ve_i$, not affecting others.

\section{Attention with Bounded Memory}\label{sec:abc}

A straightforward way to improve attention's efficiency is to bound its memory size.
Our outer-product view of attention provides a straightforward way to devise this, 
by replacing $\{\ve_i\}$ with control vectors that select $n \ll N$ vectors to attend to.
We dub this approach {\bf a}ttention with {\bf b}ounded-memory {\bf c}ontrol (\model).
Concretely, let $\widetilde{\mK},\widetilde{\mV}\in\sR^{n\times d}$ denote a constant-size memory with $n$ slots,
with $n$ set \emph{a priori}.
\begin{align}\label{eq:abc}
\widetilde{\mK}=\sum_{i=1}^{N}\vphi_{i}\otimes \vk_{i}, \quad \widetilde{\mV}=\sum_{i=1}^{N}\vphi_{i}\otimes \vv_{i}.
\end{align}
$\{\vphi_i\in\sR^{n}\}_{i=1}^{N}$ denotes a sequence of \term{control} vectors.
The output is calculated by attending to 
$\widetilde{\mK}$ and $\widetilde{\mV}$:
$\model\left(\vq, \{\vk_i\}, \{\vv_i\}, \{\vphi_i\} \right)=$
\begin{align}\label{eq:abc_softmax}
\widetilde{\mV}^\top\softmax\left(\widetilde{\mK}\vq\right).
\end{align}%
We will discuss various ways to construct $\{\vphi_i\}$
in the subsequent sections.
Reading from the memory takes a constant $\gO(n)$ time and space;
therefore \model's overall complexity is $\gO(Nn)$, linear in the sequence length.\footnote{
    Using bounded memory distinguishes \model from softmax attention.
    If growing-size memory \emph{were} allowed ($n=N$), an \model with $\vphi_i=\ve_i$ would fall back
    to softmax attention.
}

Eq.~\ref{eq:abc} offers an equivalent recurrent computation,
which is particularly useful in causal attention where only the prefix is looked at, %as we will explore in our experiments.
\begin{align}\label{eq:abc_recurrent}
    \widetilde{\mK}_{t+1} = \widetilde{\mK}_{t} + \vphi_{t+1}\otimes \vk_{t+1},  % \quad 
    % \widetilde{\mV}_{t+1} =\widetilde{\mV}_{t} + \vphi_{t+1}\otimes \vv_{t+1}.
\end{align}
likewise for $\widetilde{\mV}_t$.
$\widetilde{\mK}_t$ and $\widetilde{\mV}_t$ can be seen as the \term{recurrent hidden state} that
encodes the prefix. % $\vx_{\leq t}$.

In what follows, we study several existing efficient attention approaches and show that 
they are in fact instances of the \model abstraction.

\subsection{Linformer} \label{sec:linformer}
Linformer~\citep{wang2020linformer} is an established efficient transformer variant that has proven successful in masked language modeling and text encoding.
It assumes fixed-length inputs and
learns a low-rank approximation of the attention weights.
A learned $n$-by-$N$ matrix $\mW^{\text{LF}}$ down projects the $N$-by-$d$ dimensional 
keys and values along the \emph{timestep} dimension, to an $n$-by-$d$ memory: 
$\widetilde{\mK}^{\text{LF}}=\mW^{\text{LF}}\mK$, $\widetilde{\mV}^{\text{LF}}=\mW^{\text{LF}}\mV$;
they are then used for attention computation with~Eq.~\ref{eq:abc_softmax}.
This yields a linear complexity in the input length.
Linformer is an \model instance with $\vphi_i^{\text{LF}}=\mW_{:,i}^{\text{LF}}$ ($i$th column),
and in this sense, it learns a control vector for each position.

Previous works have noted that Linformer \emph{cannot} be efficiently applied in causal attention~(Table 1 of \citealp{tay2020}). 
Indeed, it is less straightforward to avoid mixing future with the past when projecting along the timestep dimension.
\model reveals that, in fact,
Linformer \emph{is} applicable in causal attention.
Like all \model models, it admits 
a linear-complexity recurrent computation (Eq.~\ref{eq:abc_recurrent}):
$ \widetilde{\mK}_{t+1}^{\text{LF}} = \widetilde{\mK}_{t} + \vphi_{t+1}^{\text{LF}}\otimes \vk_{t+1}$.
This confirms \model's benefits: it reveals new insights about existing models
and reassesses their applications and impact.
Our experiments show that Linformer achieves competitive performance in machine translation.

\subsection{Clustering-Based Attention}\label{sec:cluster}
Improving attention's efficiency with clustering has received an increasing amount of interest~\interalia{kitaev2020reformer,roy2020efficient,wang2020cluster}.
\model bears interesting connections to clustering-based methods.
Here we discuss an approach that closely follows \citet{vyas2020fast}, 
except that it clusters keys and values instead of queries,
and only attends to the centroids
to reduce the effective context size.
% it clusters keys/values,
% and only attends to the centroids.
Formally, keys and values are grouped into $n<N$ clusters $\{\widetilde{\vk}^{\text{CL}}_j\}_{j=1}^{n}$,
$\{\widetilde{\vv}^{\text{CL}}_j\}_{j=1}^{n}$.\footnote{
We use $\widetilde{\vk}^{\text{CL}}_j$
to denote both the $j$th cluster and its centroid.
% Likewise for the values.
}
Let an $N$-by-$n$ binary matrix $\mM$ denote the cluster membership shared between keys and values.
$M_{i,j}=1$ iff.\ $\vk_i$ is assigned to cluster $\widetilde{\vk}^{\text{CL}}_j$
and $\vv_i$ to $\widetilde{\vv}^{\text{CL}}_j$.
The $j$th centroid for the keys is % and values are respectively
\begin{align}
    \widetilde{\vk}^{\text{CL}}_j = \sum_{i=1}^{N}\frac{M_{i,j}}{\sum_{\ell=1}^{N}M_{\ell,j}}\vk_i;
    % \quad\widetilde{\vv}^{\text{CL}}_j= \sum_{i=1}^{N}\frac{M_{i,j}}{\sum_{\ell=1}^{N}M_{\ell,j}}\vv_i.
\end{align}
likewise for the values.
It then attends over the centroids using Eq.~\ref{eq:abc_softmax},
%over $\{\widetilde{\vk}^{\text{CL}}_j\}_{j=1}^{n}$ and $\{\widetilde{\vv}^{\text{CL}}_j\}_{j=1}^{n}$:
%$(\widetilde{\mV}^{\text{CL}})^\top\softmax(\widetilde{\mK}^{\text{CL}}\vq)$, where
with $\widetilde{\mK}^{\text{CL}}=[\widetilde{\vk}^{\text{CL}}_1, \dots, \widetilde{\vk}^{\text{CL}}_n]^\top=$
\begin{align*}
\begin{split}
    % \widetilde{\mK}^{\text{CL}}=[\widetilde{\vk}^{\text{CL}}_1, \dots, \widetilde{\vk}^{\text{CL}}_n]^\top
    \sum_{j=1}^{n}\ve_j \otimes \widetilde{\vk}^{\text{CL}}_j
    &= \sum_{j=1}^{n}\ve_j \otimes \sum_{i=1}^{N}\frac{M_{i,j}}{\sum_{\ell=1}^{N}M_{\ell,j}}\vk_i\\
    &= \sum_{i=1}^{N}\left(\sum_{j=1}^{n}\ve_j \frac{M_{i,j}}{\sum_{\ell=1}^{N}M_{\ell,j}}\right)\otimes\vk_i.
\end{split}
\end{align*}
The last line indicates that this model is an instance of \model:
$\vphi_i=\sum_{j=1}^{n}(M_{i,j}/\sum_{\ell=1}^{N}M_{\ell,j}) \ve_j$.
The stack of centroids can be seen as the constant-size memory.
% The above approach closely follows \citet{vyas2020fast}, except that it clusters keys and values instead of queries.
Putting aside the clustering overhead (i.e., constructing $\mM$ and computing centroids),
it has a linear complexity in the sequence length.

\subsection{Sliding-Window Attention}\label{sec:sw}
In some applications, being able to remove entries from the memory can be beneficial:
clearing up older context frees
slots for more recent ones, promoting a locality inductive bias.
\model offers the capability to do so, if augmented with an additional matrix multiplication.
We use the sliding-window attention as an example.

Attending to the most recent $n$ input tokens~\interalia{beltagy2020longformer,zaheer2020big,sukhbaatar2021not} can be seen as a first-in-first-out queue that ``pops'' out the oldest token while ``pushing'' in the most recent one:
$\widetilde{\mK}_t^{\text{WD}}=[\vk_{t-n+1},...,\vk_t]^\top$.
The pop operation can be achieved by multiplying an $n$-by-$n$ \term{upper shift matrix}: $U_{i,j}=\delta_{i+1,j}$,
with $\delta$ being the Kronecker delta (i.e., $\mU$ has ones only on the superdiagonal and zeros elsewhere).
Left-multiplying $\mU$ against $\widetilde{\mK}_{t}^{\text{WD}}$ shifts its rows one position up, with zeros appearing in the last:
\begin{align*}
\mU \widetilde{\mK}_{t}^{\text{WD}}
&=\mU\big[\underbrace{\vk_{t-n+1}, \dots, \vk_{t}}_{n}\big]^\top\\
&=\big[\underbrace{\vk_{t-n+2}, \dots, \vk_{t-1}, \vk_{t}}_{n-1},\vzero\big]^\top\in \sR^{n\times d}.
\end{align*}
Then the most recent token can be put into the slot freed up:
% \begin{align}
    $\widetilde{\mK}_{t+1}^{\text{WD}}=\mU\widetilde{\mK}_{t}^{\text{WD}} + \ve_n\otimes \vk_{t+1}$.
% \end{align}
$\mU$ and $\vphi_t=\ve_n$ ensure a first-in-first-out queue.
Dilated and stride convolution patterns~\citep{beltagy2020longformer} can be similarly recovered (\S\ref{appx:dilated}).

Recurrently multiplying $\mU$ simulates the discrete pop operation~\citep{grefenstette2015learning,joulin2015inferring,yogatama2018memory} in a differentiable way.
This is reminiscent of recurrent neural networks,
while in this case $\mU$ is \emph{never} updated as parameters.
It is exciting to explore learning $\mU$, but is beyond the scope of this work.

\paragraph{Discussion.}

Besides the models discussed above, certain
variants of \citet{Rae2020Compressive} and sparse attention patterns~(local-to-global attention;~\citealp{beltagy2020longformer,zaheer2020big,ainslie2020etc})
can also be seen as instances of \model~(\S\ref{appx:other_abc}).
\model provides a unified perspective of them,
and at the same time points out their limitations:
their control strategies are context-agnostic.
In response to this, in \S\ref{sec:learned_phi} we propose to learn a contextualized strategy from data.
Table~\ref{tab:abc} analyzes various \model models,
and Table~\ref{tab:complexity} details their complexity.

\begin{table*}[t]
\centering
\begin{tabular}{@{}l @{\hspace{8pt}} l l @{\hspace{8pt}} l @{}}
\toprule[.1em]
\textbf{Model} 
& \textbf{Section}
 & $\vphi_t$ & \textbf{Mem. Control} \\
\midrule[.1em]
Sliding-window & \S\ref{sec:sw}  & $\ve_n$ & $\widetilde{\mK}_{t+1}=\mU\widetilde{\mK}_{t} + \vphi_{t+1}\otimes \vk_{t+1}$\\
\midrule
Linformer & \S\ref{sec:linformer} & $\mW_{:,t}^{\text{LF}}$ & \multirow{6}{*}{$\widetilde{\mK}_{t+1}=\widetilde{\mK}_{t} + \vphi_{t+1}\otimes \vk_{t+1}$} \\
L2G Pattern & \S\ref{appx:l2g} & $\ve_i$ if $\vx_t$ is the $i$th global token  & \\
\modelrandom & \S\ref{appx:random} & $\ve_{i_t}$, where $i_t\sim\operatorname{unif}\{1, n\}$\\
Comp. Trans. & \S\ref{appx:compressive} & $\ve_{\floor{nt/N}}$ &\\
Clustering & \S\ref{sec:cluster} & $\sum_{j=1}^{n}\left(M_{t,j}/\sum_{\ell=1}^{N}M_{\ell,j}\right) \ve_j$ \\
\modelmlp & \S\ref{sec:learned_phi} & $\vexp(\mW_{\vphi}\vx_t) / \sum_{i=1}^{t}\vexp(\mW_{\vphi}\vx_t)$  &\\

\bottomrule[.1em]
\end{tabular}
\caption{A comparison of different \model models.
$N$ denotes the sequence length, and $n$ the memory size. 
$\vphi_t$ denotes the memory control vector for $\vk_t$ and $\vv_t$, and $\operatorname{unif}$ is the discrete uniform distribution.
\resolved{\jungo{t->i for consistency with main text?}} } 
\label{tab:abc}
\end{table*}

\begin{table*}[t!]
\centering

    \begin{tabular}{@{} l ccc m{0.001em} ccc@{}}
    \toprule[.1em]
    & \multicolumn{3}{c}{\textbf{Time Complexity}}
    & 
    & \multicolumn{3}{c}{\textbf{Space Complexity}}\\
    \cmidrule(lr){2-4}
    \cmidrule(lr){6-8}
    \textbf{Model}  
    & \textbf{Mem.} & \textbf{Per Query} & \textbf{Overall}
    &
    & \textbf{Mem.} & \textbf{Per Query} & \textbf{Overall}\\
    \midrule
    Softmax Attention
    & - & $\mathcal{O}(N)$ & $\mathcal{O}(N^2)$
    & 
    & - & $\mathcal{O}(N)$ & $\mathcal{O}(N^2)$\\
    \midrule
    \model
    & $\mathcal{O}(N)$ & $\mathcal{O}(n)$ & $\mathcal{O}(nN)$
    & 
    & $\mathcal{O}(n)$ & $\mathcal{O}(n)$ & $\mathcal{O}(nN)$\\
    \bottomrule[.1em]
    \end{tabular}

\caption{\model's time and space complexity in sequence length against the softmax attention's.
``Mem.'' indicates the time and space needed for calculating and storing memory $\widetilde{\mK}, \widetilde{\mV}$.
$N$ denotes the sequence length, and $n$ the memory size.
The time complexity analysis assumes that the softmax attention \emph{cannot} be parallelized across the queries.
In practice, this is common in autoregressive decoding or for long sequences where 
the accelerators (e.g., GPUs) do not have enough threads
to fully parallelize softmax attention's computation across different queries.
}
\label{tab:complexity}
\end{table*}

\section{Learned Memory Control}\label{sec:learned_phi}

The \model abstraction connects several existing approaches
that would otherwise seem distinct.
This inspires the design of new architectures.
We hypothesize that learning a contextualized strategy can
achieve better performance.
This section introduces \modelmlp. It parameterizes $\vphi$ with a single-layer multi-layer perceptron (MLP)
that takes as input the token's representation $\vx_i$, and determines which slots to write it into and how much.
\begin{align}\label{eq:phi_mlp}
    \valpha_i = \vexp\left(\mW_{\vphi}\vx_i\right),\quad
    \vphi_i = \valpha_i \left/ \sum_{j=1}^{N}\valpha_j.\right.
\end{align}
Matrix $\mW_{\vphi}$ is learned. 
$\vexp$ is an elementwise activation function.
The motivation is to allow for storing a ``fractional'' (but \emph{never} negative) amount of input into the memory.\footnote{
We experiment with other activations in \S\ref{appx:mt}.
\resolved{\jungo{concise summary? improve or not?}\hao{don't have the space}}}
Using a non-negative activation, however, has a drawback:
the scales of $\sum_i \vphi_i\otimes \vk_i$ and $\sum_i \vphi_i\otimes \vv_i$
would grow with the sequence lengths, making training less stable.
To overcome this, we divide $\valpha_i$ vectors by their sum.
This functions as normalization and aims to offset the impact of varying sequence lengths.\footnote{
Here encoder self-attention or cross attention is assumed, and the normalization sums over the entire sequence.
\rev{Causal attention is slightly different, normalizing by the sum over the prefix instead: $\vphi_i = \valpha_i /\sum_{j=1}^{i}\valpha_j$.
This does \emph{not} require access to future tokens.
\S\ref{appx:causal_abc} details a linear complexity computation graph of causal $\vphi_i$.}
}
It admits the recurrent computation graph as in Eq.~\ref{eq:abc_recurrent}, and has a linear complexity in the sequence length.

\rev{
A key design choice of \modelmlp is that its $\vphi_i$ depends \emph{only} on
current input $\vx_i$.
This helps (1) keep the recurrent computation efficient in practice~\citep{lei2018sru},
and (2) make it applicable in not only encoder self-attention and cross attention,
but also causal attention.
Concurrently to this work, \citet{goyal2021sw} and \citet{ma2021luna} also proposed methods to learn contextualized control.
They compute $\vphi_i$ from \emph{previous} layer's memory,
revealing the full sequence to the control vectors.
As a result, these two approaches are \emph{unsuitable} for causal attention.\footnote{
\rev{Both are instances of \model (\S\ref{appx:luna}).
%See \S\ref{appx:luna} for a detailed discussion.
\citet{ma2021luna} resorts to a variant of \citet{katharopoulos20transformers}
for causal attention.}
}
}

\modelmlp, as other \model models, can be used as a drop-in replacement for the canonical softmax attention, and
we apply its multihead variant in transformers.
With proper parameter sharing, the number of additional parameters \modelmlp incurs is small:
inspired by \citet{wang2020linformer}, we tie $\vphi$-MLP's parameters across different layers,
which adds less than 1\% parameters to the models.

\paragraph{\modelmlp: context-agnostic then context-dependent attention.}
We now dissect \modelmlp and show that it can be seen as
a cascade of two attention mechanisms:
one with a learned context-agnostic ``pseudo query'' followed by one with a context-dependent query.
Our analysis starts with a one-dimensional example; the conclusion generalizes to higher-dimensional cases.
\begin{example}\label{example:exp_phi}
Consider \modelmlp with a \emph{single} memory slot ($n=1$). It is parameterized with a learned vector $\vw_\phi$,
and $\phi_i = \exp(\vw_{\phi}\cdot \vx_i) / \sum_{j=1}^{N}\exp(\vw_{\phi}\cdot \vx_j)$.
Since $\phi_i$ is a scalar here, $\phi_i\otimes\vk_i=\phi_i\vk_i^\top$.
\begin{align*}
    \widetilde{\mK}^\top
    &= \sum_{i=1}^{N}\left(\phi_i\otimes\vk_i\right)^\top\\
    &=\sum_{i=1}^{N}\frac{\exp(\vw_\phi\cdot\vx_i)}{\sum_{j=1}^{N}\exp(\vw_\phi\cdot\vx_j)}\vk_i\\
    &=\attn\left(\vw_\phi,\{\vx_i\}_{i=1}^{N}, \{\vk_i\}_{i=1}^{N}\right).
\end{align*}
\end{example}
In other words, $\widetilde{\mK}$ uses $\vw_\phi$ as a ``pseudo-query'' to attend to
$\{\vx_i\}$ and $\{\vk_i\}$. 
Likewise, $\widetilde{\mV}^\top=\attn(\vw_\phi,\{\vx_i\}_{i=1}^{N}, \{\vv_i\}_{i=1}^{N})$.
Despite its similarity to the standard softmax attention,
Example~\ref{example:exp_phi} has a more efficient linear complexity in sequence lengths.
$\vw_\phi$'s being context-independent is the key to the savings.
Table~\ref{tab:complexity} details its complexity.  % to softmax attention's.

Example~\ref{example:exp_phi}'s conclusion generalizes to higher-dimensional cases:
the $j$th dimension of $\{\vphi_i\}$ attends to $\{\vx_i\}$ and $\{\vk_i\}$ using the $j$th row of $\mW_{\vphi}$ as the context-independent pseudo-query;
$n$ such attention mechanisms run in parallel, stacking the results into $n$-by-$d$ memory
$\widetilde{\mK}$ and $\widetilde{\mV}$.
Intuitively, it is the ``real queries'' $\{\vq_i\}$ that encode ``what information is useful for the prediction task.''
Without access to them, \modelmlp summarizes the input for $n$ times using different pseudo-queries,
aiming to preserve enough information in the memory for onward computation. 
The attention output is calculated with the context-dependent real queries using Eq.~\ref{eq:abc_softmax}.
\S\ref{appx:high_dimensional} presents a detailed derivation.

\paragraph{Connections to other prior works.}
Although starting from distinct motivations, \modelmlp closely relates to hierarchical attention (HA;~\citealp{yang2016hierarchical}).
HA summarizes the context into higher-level representations with a cascade of attention mechanisms, e.g.,
words to sentences, and then to documents.
\modelmlp applies two types of attention. 
The first learns context-agnostic pseudo-queries 
and attends to the same sequence for $n$ times in parallel, 
while the second retrieves from the memory with real queries.
HA, in contrast, summarizes non-overlapping segments at each level. 

\rev{
The learned pseudo-queries closely relate to the inducing point method in set attention~(ISA;~\citealp{lee2019set}).
ISA applies a non-linear feedforward network between a cascade of two attention modules.
This precludes the outer-product memory computation and efficient recurrences in \model.
}

Another line of work ``linearizes'' attention through kernel tricks
and also applies bounded memory:
their feature map dimensions are analogous to memory sizes.
They substitute 
the softmax with approximations~\citep{peng2021rfa,choromanski2020masked},
heuristically designed~\citep{katharopoulos20transformers,schlag2021linear}, or learned~\citep{kasai2021t2r} functions.
\modelmlp keeps the softmax, but over a smaller constant-sized context.
This can be useful in practice:
(1) \model provides a unified perspective of several efficient attention methods, allowing for borrowing from existing wisdom to design new architectures;
(2) it draws a close analogy to the canonical softmax attention,
and is better-suited as its drop-in substitute in various application settings, as we will show in the experiments;
(3) empirically, we find that \modelmlp can get away with a much smaller memory size to retain the accuracy.
\citet{peng2021rfa} and \citet{schlag2021linear} use gating to promote recency bias.
The same technique is equally applicable in \model models.

\rev{
The learned contextualized memory control is reminiscent of the content-based addressing in neural Turing machines~(NTM;~\citealp{graves2014neural}).
\modelmlp computes the control vectors $\{\vphi_i\}$ as a function of the input,
but \emph{not} of the memory as in NTM.
This ensures that the control vectors at different timesteps can be computed in parallel, 
improving the time efficiency in practice~\citep{lei2018sru,peng2018rational}.
Analogies between memory and neural architectures
are also made by other previous works~\interalia{hochreiter1997lstm,weston2014memory,le2020self}.
}

\section{Experiments}\label{sec:experiments}
We evaluate \model models on language modeling (\S\ref{sec:lm}), sentence-level and document-level machine translation (\S\ref{sec:mt}),
and masked language model finetuning (\S\ref{sec:mlm}).
Dataset statistics and implementation details are summarized in \S\ref{appx:experiments}.
\subsection{Language Modeling}\label{sec:lm}

\paragraph{Setting.}
We experiment with WikiText-103, sampled text from English Wikipedia~\citep{merity2016pointer}.
The \base model with standard softmax attention is the strong transformer-based language model by~\citet{baevski2018adaptive}. 
We compare the following \model variants, which build on \base, but replace the softmax attention with linear-complexity 
bounded-memory attention alternatives
while keeping other components the same.
\begin{itemize}[nosep,leftmargin=1em,labelwidth=*,align=left]
%\begin{compactitem}
\item \modelmlp, as described in \S\ref{sec:learned_phi}, learns a contextualized $\exp$-MLP as the $\vphi$ function.
\item Linformer~(\S\ref{sec:linformer}; \citealp{wang2020linformer}). %, as described in \S\ref{sec:linformer}.
\item \modelrandom stores each token in a randomly-selected memory slot with $\vphi_t=\ve_{i_t}$. $i_t$ is uniformly
drawn from $\{1,\dots,n\}$ at each time step.
This helps us quantify the differences between random and learned bounded-memory controls.
\end{itemize}
We consider two model size settings:
\begin{itemize}[nosep,leftmargin=1em,labelwidth=*,align=left]
\item
16 layers \citep{baevski2018adaptive}.
All models have around $\sim$242M parameters.
They train with 512-token segments,
and evaluate with 0 or 480 context sizes:
a 0- or 480- length prefix precedes each evaluation segment.
\item 32 layers \citep{kasai2021t2r}.
All models have  $\sim$484M parameters.
This setting applies layer dropout~\citep{fan2020reducing}, 
and evaluates with a 256 context size.
It aims to compare \modelmlp to several kernel-based efficient attention variants:
ELU~\citep{katharopoulos20transformers},
RFA~\citep{peng2021rfa},
and T2R~\citep{kasai2021t2r}.
\end{itemize}

\begin{table}[t]
% \begin{table*}[th]
\centering
\begin{subtable}[b]{.48\textwidth}
\centering
\begin{tabular}{@{} l c cc m{0.01em} cc@{}}
\toprule[.1em]

&& \multicolumn{2}{c}{\textbf{Dev.}}
&& \multicolumn{2}{c}{\textbf{Test}}\\
\cmidrule(lr){3-4}
\cmidrule(lr){6-7}
\textbf{Model} 
& \textbf{$n$}
& \textbf{0} & \textbf{480}
&& \textbf{0} & \textbf{480}\\

\midrule[.1em]
\base & - & 19.8 & 18.4 && 20.5 & 19.0\\

\midrule

Linformer & 64 & 26.5 & 27.1 && 27.2 &30.7\\ 
\modelrandom & 64 & 23.2 & 22.3 && 24.0 & 23.1\\ 

\midrule
\modelmlp & 32 & 21.2 & 19.7 && 21.9 & 20.5 \\ 
\modelmlp & 64 & \textbf{20.4} & \textbf{18.9} && \textbf{21.1} & \textbf{19.5}\\ 

\bottomrule[.1em]
\end{tabular}
\caption{\label{tab:lm16}16-layer setting.
0/480 indicate evaluation context sizes.
} 
\end{subtable}%
\hfill
\vspace{.25cm}
\begin{subtable}[b]{.48\textwidth}
\centering
\begin{tabular}{@{} l c cc @{}}
\toprule[.1em]

\textbf{Model} 
& \textbf{$n$}
& \textbf{Dev.} & \textbf{Test}\\

\midrule[.1em]
$\dagger$\base & - & 17.9 & 18.5\\

\midrule

$\dagger$ELU & 128 & 22.0 & 22.8 \\
$\dagger$RFA & 32 & 20.4 & 21.3 \\
$\dagger$T2R & 32 & 20.1 & 20.8 \\

\midrule
\modelmlp & 32 & {\bf 19.2} & {\bf 19.9}\\ 

\bottomrule[.1em]
\end{tabular}
\caption{\label{tab:lm32}32-layer setting.
A 256-length context is used at evaluation time.
$\dagger$ numbers are due to \citet{kasai2021t2r}.
} 
\end{subtable}
\caption{WikiText-103 language modeling perplexity (\textbf{lower is better}).
$n$ denotes the memory size.
Bold numbers perform the best among linear-complexity models.
}
%\vspace{-.25cm}
\end{table}

\paragraph{Results.}
Table~\ref{tab:lm16} compares \model variants using \citet{baevski2018adaptive}'s 16-layer setting.
Among \model models, \modelmlp achieves the best performance for both context sizes.
With a memory size $n=64$, \modelmlp outperforms both Linformer and \modelrandom
by more than 2.9 test perplexity; and the gap is larger with the longer 480-length context:
more than 3.6 test perplexity.
\modelmlp-32 outperforms
its larger-memory \model counterparts by more than 2.1 test perplexity.
These results confirm \modelmlp's advantages of using a contextualized strategy.
Surprisingly, Linformer \emph{underperforms} \modelrandom,
\rev{and its performance drops with the larger 480-length context window.}
This suggests that, while successful in text encoding,
Linformer's position-based strategy is a suboptimal design choice for causal attention\rev{, at least for long context}.
All \model models \emph{underperform} the \base,
with \modelmlp-64 having the smallest gap of $0.5$ perplexity.
\modelmlp-32 outperforms kernel-based methods by more than 0.9 test perplexity,
using \citet{kasai2021t2r}'s 32-layer setting (Table~\ref{tab:lm32}).

\begin{table}[t]
\centering
\begin{subtable}[b]{.47\textwidth}
\centering
\begin{tabular}{@{} l  cc c@{}}
\toprule[.1em]
\textbf{Model} 
& \textbf{Cross $n$}
& \textbf{Causal $n$}
& \textbf{BLEU}\\

\midrule[.1em]
\base & - & - & 27.2 \\
\midrule
\modelrandom & 32 & 32 & 25.7 \\
\modelrandom & 64 & 64 & 26.2 \\
\midrule
Linformer & 32 & 32 & \rev{26.6}\\
Linformer & 64 & 64 & \rev{26.7}\\
\midrule
\modelmlp & 32 & \phantom{0}8 & 27.1\\
\modelmlp & 32 & 32 & \textbf{27.3}\\

\bottomrule[.1em]
\end{tabular}
\caption{\label{tab:wmt}%WMT14 EN-DE sentence-level machine translation results.
Bolded number outperforms \base.
%All models have around 61M parameters.
} 
\end{subtable}%
\hfill
\vspace{.25cm}
\begin{subtable}[b]{.47\textwidth}
\centering
\begin{tabular}{@{} l  cc c@{}}
\toprule[.1em]
\textbf{Model} 
& \textbf{Cross $n$}
& \textbf{Causal $n$}
& \textbf{BLEU}\\

\midrule[.1em]
\base & - & - & 39.9 \\
\midrule
Linformer & 128 & 64 &  -\\
\midrule
\modelrandom & 128 & 64 & 38.6 \\
\midrule
\modelmlp & 128 & 64 &  \textbf{39.7}\\

\bottomrule[.1em]
\end{tabular}
\caption{\label{tab:iwslt}% IWSLT14 ES-EN document-level machine translation results.
Linformer fails to converge even with multiple random seeds.
Bold number performs the best among \model models.
} 
\end{subtable}
\caption{Machine translation test SacreBLEU.
Left: sentence-level translation with WMT14 EN-DE;
right: document-level translation with IWSLT14 ES-EN.
}
\end{table}

\subsection{Machine Translation}\label{sec:mt}
\paragraph{Datasets.} 
To assess their performance over various output lengths,
we compare \model models on sentence- and document- level machine translation. 

\begin{itemize}[nosep,leftmargin=1em,labelwidth=*,align=left]
\item Sentence-level translation with WMT14 EN-DE \citep{bojar2014wmt}.
The preprocessing and data splits follow \citet{vaswani2017attention}.
\item Document-level translation with IWSLT14 ES-EN \citep{cettolo2014report}.
We use \citet{miculicich2018document}'s data splits and preprocessing.
Following standard practice~\citep{voita2019good}, a 4-sentence sliding window is used to
create the dataset, i.e., each instance has 4 sentences.
\end{itemize}

\paragraph{Setting.}
We compare \model variants as in~\S\ref{sec:lm}.
\S\ref{appx:mt} further compares to the clustering-based (\S\ref{sec:cluster})
and sliding-window (\S\ref{sec:sw}) \model variants.

% We compare the \model variants described in~\S\ref{sec:lm}.
The \base model they build on is our implementation of transformer-base~\citep{vaswani2017attention}. 
\model variants replace decoder cross attention and causal attention with bounded-memory attention,
while keeping softmax attention for the encoder, since its overhead is much less significant~\citep{kasai2020deep};
other components are kept the same.
\S\ref{appx:mt} studies a model
that replaces \emph{all} softmax attention with \modelmlp.
It performs on par with \base, confirming \modelmlp's broad applicability 
in various application scenarios. 
We evaluate with SacreBLEU \citep{matt2014call}.
% Further details are described in \S\ref{appx:experiments}.

\paragraph{Results.}
Table~\ref{tab:wmt} summarizes sentence-level machine translation
results on the WMT14 EN-DE test set.
Overall \modelmlp performs on par with \base,
with either 32-32 cross-causal memory sizes or 32-8.
Even with smaller memory sizes, it outperforms other \model variants by more than 1.1 BLEU.
Differently from the trend in the language modeling experiment~(\S\ref{sec:lm}),
Linformer outperforms \modelrandom by more than 0.5 BLEU.
We attribute this to the smaller sequence lengths of this dataset.
\modelmlp outperforms other \model models by more than 0.4 BLEU,
even with smaller memory sizes.

The trend is similar on document-level translation with IWSLT14 ES-EN (Table~\ref{tab:iwslt}),
except that \modelmlp slightly \emph{underperforms} \base by 0.2 BLEU.
This suggests that even with longer sequences, \modelmlp is effective despite its bounded memory size.
\rev{Linformer fails to converge even with multiple random seeds,
suggesting the limitations of its purely position-based strategy in 
tasks involving decoding varying-length text.}

\subsection{Masked Language Model Finetuning} \label{sec:mlm}
\paragraph{Setting.}
We compare the \model variants as in~\S\ref{sec:lm}. % on masked language modeling.
It is interesting to pretrain \model from scratch,
but we lack the resources to do so.
Instead, we warm-start from a pretrained RoBERTa-base~\citep{liu2019roberta} trained with the softmax transformer,
swap its attention with \model variants, 
and continue pretraining with the masked language modeling (MLM) objective 
on a concatenation of 
BookCorpus~\citep{zhu2015aligning},
English Wikipedia,
OpenWebText~\citep{gokaslan2019open},
and RealNews~\citep{zellers2019defending}.\footnote{
Our data differs from RoBERTa's, which we do \emph{not} have access to.
We replace CC-News~\citep{nagel2016news} with RealNews, and drop Stories~\citep{trinh2018simple},
whose public access is broken 
at the time of this work. %, the public access to the Stories dataset is broken.
%: \url{https://console.cloud.google.com/storage/browser/commonsense-reasoning/reproduce/stories_corpus?pli=1}.
}
Then the models are finetuned and evaluated on downstream classification datasets from the the GLUE benchbark~\citep{wang2018glue}.
This is an appealing setting, 
since it avoids reinvesting the huge amounts  of  resources  already  put  into  pretraining.\footnote{
In preliminary experiments, we explored swapping in \model, and then directly finetuning on downstream tasks \emph{without} continued MLM pretraining;
all models fail.
}

\paragraph{Results.}
Table~\ref{tab:glue} compares downstream text classification performance.
\base indicates a baseline that continues pretraining RoBERTa-base on our data.\footnote{
\base slightly \emph{underperforms} RoBERTa-base.
This could be due to overfitting, or the pretraining data discrepancy. %a discrepancy between the datasets. 
}
Following standard practice, we report development accuracy.
Linformer achieves competitive performance, aligned with \citet{wang2020linformer}'s results.
\modelmlp outperforms Linformer, and performs on par with or better than \base,
affirming the benefits of using contextualized memory organization in MLM.
\modelrandom fails to converge in continued pretraining even with multiple seeds.

Based on the above results, we think \modelmlp can achieve competitive performance when pretrained from scratch,
just as Linformer does \citep{wang2020linformer}.
Further empirical exploration is beyond our budget and left for future work.

\begin{table}[t]
\centering
\begin{tabular}{@{} l @{\hspace{6pt}} c @{\hspace{6pt}} c @{\hspace{6pt}} c @{\hspace{6pt}}  c @{\hspace{6pt}}  c @{\hspace{6pt}}  c@{}}
\toprule[.1em]
\textbf{Model} 
& \textbf{$n$}  & \textbf{MNLI} & \textbf{QNLI} & \textbf{QQP} &\textbf{SST} & \textbf{Avg.} \\

\midrule[.1em]
\base & - & 87.2 & 92.4 & 91.7 & 94.3 & 91.4 \\ 
\midrule
Linformer & \phantom{0}64 & 85.3 & 91.8 & 90.8 & 92.4 & 90.1 \\
Linformer & 128 & 86.1 & 91.9 & 91.4 & 93.7 & 90.8 \\
\modelmlp & \phantom{0}64 & 85.6 & 91.8 & \underline{91.7}& 93.8 & 90.7 \\
\modelmlp & 128 &  \textbf{87.1} & \underline{\bf 92.6} & \underline{\bf 91.8} & \underline{\bf 94.4} & \underline{\bf 91.5} \\

\bottomrule[.1em]
\end{tabular}
\caption{\label{tab:glue}Text classification development set accuracy.
All models continue pretraining RoBERTa-base on our data with the MLM objective.
Bold numbers perform the best among \model models,
and underlined ones perform on par with or better than \base.
} 
\end{table}

\section{Analysis}\label{sec:analysis}

\paragraph{Decoding efficiency over varying sequence lengths.}
\model's efficiency gains can be more prominent for long sequences. 
We study \modelmlp's decoding overhead with varying sequence lengths.
Following \citet{kasai2021t2r}, 
we consider a sequence-to-sequence generation experiment.
Three linear-complexity models are compared: 
RFA~(with 256/128 cross/causal memory sizes; \citealp{peng2021rfa}),
T2R~(32/4; \citealp{kasai2021t2r}),
and \modelmlp~(32/8).
The sizes are chosen to
maximize efficiency \emph{without} accuracy drop.
T2R needs to be finetuned from a pretrained transformer to match its performance, while others don't.

\begin{figure}[t]%{0.48\textwidth}
\centering
\begin{subfigure}[b]{0.47\textwidth}
\centering
\includegraphics[trim={1.5cm 0cm 1.5cm 0cm},clip,width=.85\textwidth]{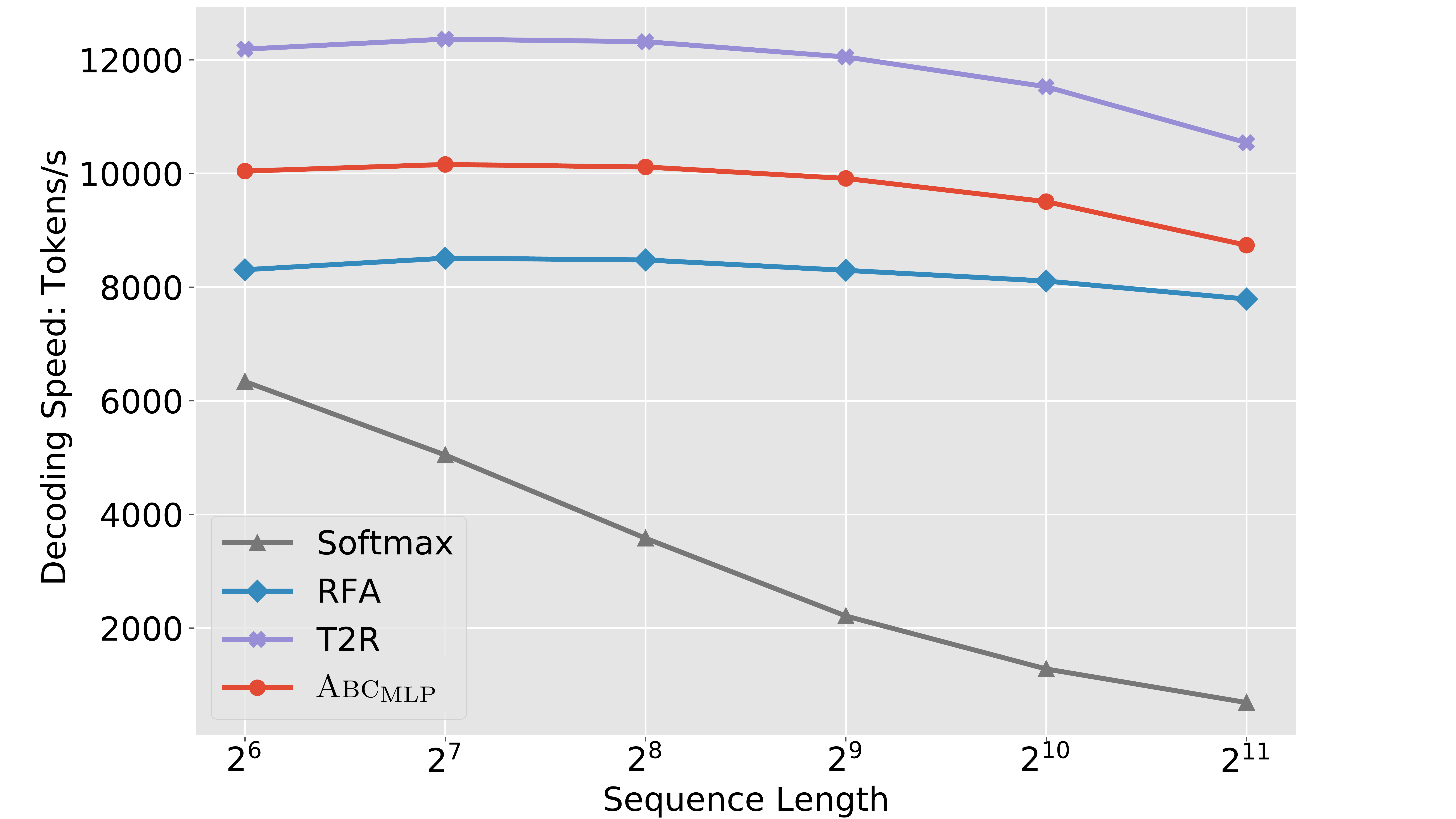}
\caption{Decoding Speed.}
\label{fig:speed_length}
\end{subfigure}%
\hfill
\vspace{.2cm}
\begin{subfigure}[b]{0.47\textwidth}
\centering
\includegraphics[trim={1.5cm 0cm 1.5cm 0cm},clip,width=.85\textwidth]{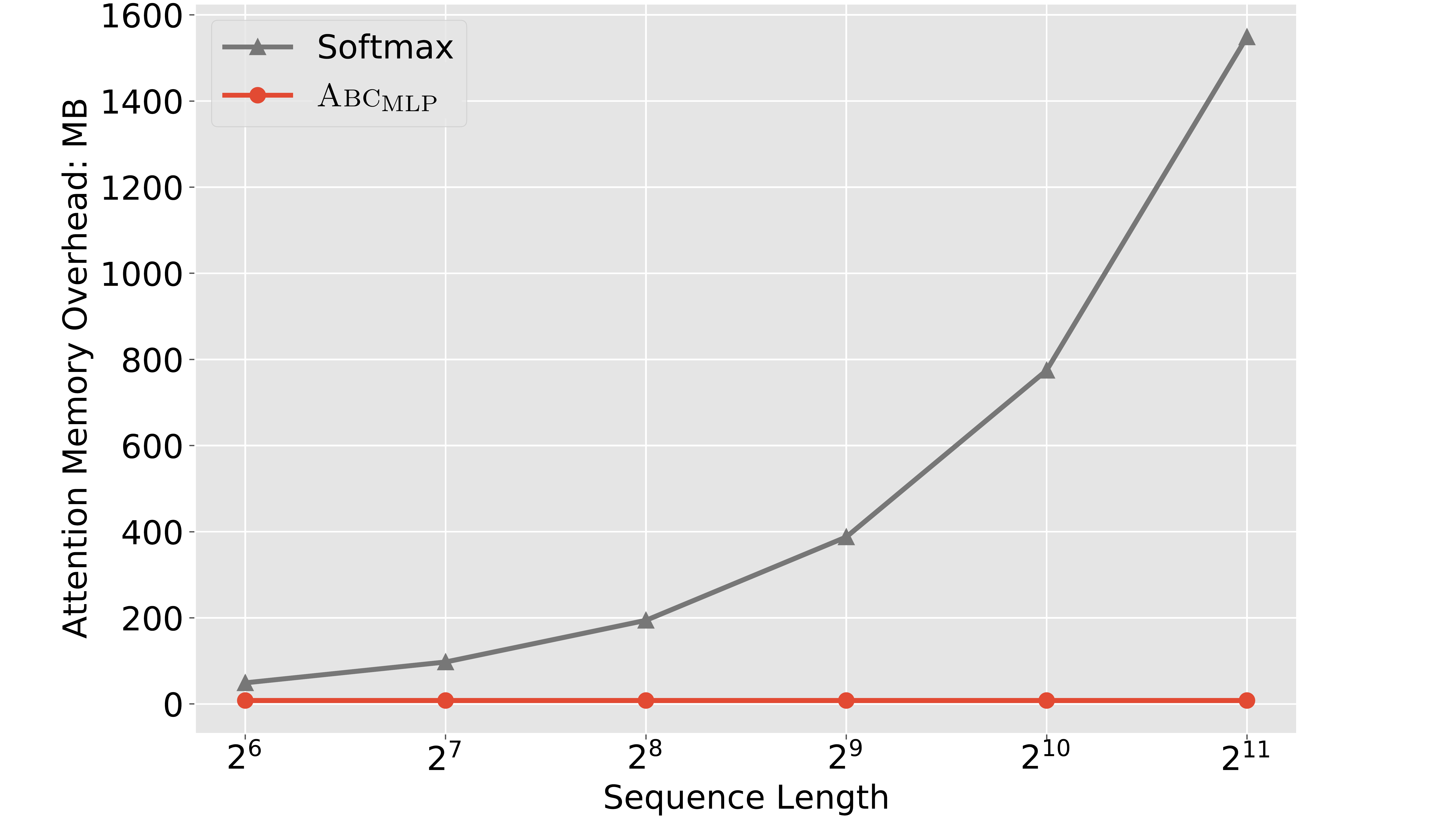}
\caption{Decoding memory overhead.}
\label{fig:memory_length}
\end{subfigure}
\hfill
\caption{Sequence-to-sequence decoding speed (top) and memory consumption (bottom) varying sequence lengths.
Greedy decoding is used, with batch size 16.}
\label{fig:efficiency}
\end{figure}

All linear-time models achieve consistent decoding speed for different lengths~(Figure~\ref{fig:speed_length}),
substantially outpacing the softmax attention baseline, especially for long sequences.
In particular, \modelmlp decodes $\sim$1.25 times faster than RFA, another competitive model
that can match transformer's accuracy \emph{without} a warm start from a pretrained model.
This can be attributed to the fact that \modelmlp achieves similar accuracy with a much smaller memory.
T2R's memory sizes are similar to \modelmlp's, but it decodes about 20\% faster.
This is because it does \emph{not} compute the softmax when calculating attention output, while
\modelmlp does~(Eq.~\ref{eq:abc_softmax}).
These results show that \modelmlp is an appealing modeling choice for decoding tasks,
especially when training from scratch is desired.

\modelmlp also achieves significant savings in terms of memory overhead (Figure~\ref{fig:memory_length}).
\modelmlp, RFA, and T2R's curves are similar.

\paragraph{Text encoding efficiency.}
We compare the efficiency of \modelmlp against softmax attention and Linformer when used as text encoders.
The models' sizes mirror those in the MLM experiment (\S\ref{sec:mlm}).
Table~\ref{tab:mlm_efficiency} summarizes inference time and memory overhead 
with 512-length inputs, batch size 16.
Both \modelmlp and Linformer achieve inference speed gains and memory savings over \base.
Linformer is faster, since its linear projection is cheaper to compute than \modelmlp's MLP.
Inference speed is measured on the same V100 GPU.
The trend in memory overhead is similar.

Although \modelmlp slightly underperforms Linformer in terms of inference speed,
it can be a more appealing architectural choice in practice:
in all of our 5 experiments, \modelmlp outperforms other \model models in accuracy.
Linformer, in contrast, fails to converge or yields sub-optimal performance on some tasks. 
This confirms its flexibility and applicability in various settings. 

\paragraph{Memory size's impact on accuracy.}
Practically, one may want to minimize the memory size to improve efficiency.
We use the WMT14 EN-DE experiment to investigate how memory size affects accuracy.
Using the \S\ref{sec:mt}'s setup,
we vary \modelmlp's cross and causal attention memory sizes and 
compare their translation quality on the development data.
They are selected from $\{8, 16, 32, 64\}$,
with cross attention's equal to or larger than causal's:
cross attention is more important than causal attention
in machine translation~\citep{michel2019are}.
Our results (Table~\ref{tab:cross_causal}) align with this observation:
when cross attention memory is large enough,
reducing causal attention memory size from 64 to 8 has a minor 0.3 BLEU drop.
Surprisingly, \modelmlp with 8-8 sized cross-causal memory
is only 1.1 BLEU behind the best-performing configuration.

\begin{table}[t]
\centering
\begin{tabular}{@{} l c cc  cc@{}}
\toprule[.1em]
& \base & \multicolumn{2}{c}{Linformer} & \multicolumn{2}{c}{\modelmlp}\\
\cmidrule(lr){3-4}
\cmidrule(lr){5-6}
$n$ & - & 64 & 128 & 64 & 128\\

\midrule[.1em]
{\bf Speed}& 1.0$\times$ & 1.7$\times$ & 1.5$\times$ & 1.5$\times$ & 1.3$\times$\\
\midrule
{\bf Memory}& 1.0$\times$ & 0.5$\times$ & 0.6$\times$ & 0.5$\times$ & 0.6$\times$ \\

\bottomrule[.1em]
\end{tabular}
\caption{\label{tab:mlm_efficiency}
Text encoding inference speed (higher is better) and memory (lower is better). 
Inputs are text segments with 512 tokens and batch size 16.
} 
\end{table}
  
\begin{table}[t]%{0.42\textwidth}
\centering
\begin{tabular}{|c|r V{3} c c c c|}
\Xhline{2\arrayrulewidth}
\multicolumn{2}{|c V{3}}{} &\multicolumn{4}{c|}{\textbf{Cross $n$}} \\
 \cline{3-6}
 \multicolumn{2}{|c V{3}}{}  & 8 & 16 & 32 & 64\\\Xhline{2\arrayrulewidth}
\multirow{4}{*}{\rotatebox{90}{\textbf{Causal $n$}}}  
& 8 & 24.7 &  25.2 &  25.6 &  25.5 \\\cline{2-6}    
& 16 & - &  25.4 &  25.7 &  25.6  \\\cline{2-6}
& 32 & - &  - &  25.7 &  25.8 \\\cline{2-6}
& 64 & - &  - &  - &  25.8  \\ \cline{2-6}
   
\Xhline{2\arrayrulewidth}
\end{tabular}
\caption{\modelmlp's SacreBLEU on WMT14 EN-DE development data varying memory sizes.
}
\label{tab:cross_causal}
\end{table}

\section{Conclusion}\label{sec:conclusion}
We presented attention with bounded-memory control (\model).
It provides a unified perspective of several recently-proposed models,
and shows that they vary in the organization of the bounded memory.
\model reveals new insights into established methods and inspires new architectures.
We proposed \modelmlp,
a particular instance of \model that learns a contextualized memory control.
On language modeling, machine translation, and masked language model finetuning,
\modelmlp outperforms previous \model models.
Compared to the strong transformer baseline,
\modelmlp achieves substantial efficiency improvements 
\rev{with no or negligible} accuracy loss.

\section*{Acknowledgments}
	We would like to thank 
    the ARK group at the University of Washington
	for their helpful feedback, 
	and the anonymous reviewers
	for their thoughtful comments.
	This work was supported in part by NSF grant 2113530 and a Google Fellowship.
	Nikolaos Pappas was supported by the Swiss National Science Foundation grant P400P2\_183911.
	
\bibliography{acl}
\bibliographystyle{acl_natbib}

\clearpage
\appendix
\begin{appendices}
\section{Other \model Models}\label{appx:other_abc}

\subsection{Sparse Local-to-global Attention}\label{appx:l2g}
It sparsifies attention pattern to reduce the number of tokens that are attended to~\interalia{beltagy2020longformer,zaheer2020big}.
\resolved{\jungo{Perhaps say what the reception field means here (instead of using that term).}}
All queries attend to a subset of $n<N$ ``global tokens,'' while ignoring others.
Therefore the effective context size is reduced to $n$.
The global tokens are usually pre-selected by positions according to some heuristics. 
Local-to-global attention is an instance of \model:
it can be recovered by letting $\vphi_t = \ve_i$ if $x_t$ is the $i$th global token ($i=1,\dots,n$), and the zero vectors for others.

\subsection{Random Memory Control}\label{appx:random}
As a baseline, \modelrandom stores each token in a randomly-selected memory slot.
 This is achieved by letting $\vphi_t=\ve_{i_t}$, where $i_t$ is uniformly
drawn from $\{1,\dots,n\}$ for each $t$.
It is designed as a baseline to \modelmlp and Linformer to
quantify the differences between random and learned bounded-memory control.

Random sparse attention patterns are explored by \citet{zaheer2020big},
where a subset of $n<N$ tokens are randomly selected to be attended to by all tokens.
\modelrandom is different, and it attends to \emph{all} tokens, but randomly ``squash'' them
into an $n$-slot memory.

\subsection{Compressive Transformer with Mean Pooling}\label{appx:compressive}
The compressive transformer \citep{Rae2020Compressive} explores various ways to ``squash'' long context into
smaller and more compact representations. 
It achieves state-of-the-art performance on several language modeling benchmarks.
We show that at least the mean-pooling variant of the compressive transformer
can be seen as an \model instance.

The mean-pooling variant of the compressive transformer compresses the context by
\begin{align*}
    \mK=\big[&\vk_1, \dots, \vk_N\big]^\top\in\sR^{N\times d}\\
    \rightarrow
    \widetilde{\mK}=\bigl[
    &\underbrace{(\vk_1+\dots+\vk_c)}_{c}/c, \\
    &\underbrace{(\vk_{c+1}+\dots+\vk_{2c})}_{c}/c \dots, \\
    &\underbrace{(\vk_{N-c+1}+\dots+\vk_{N})}_{c}/c\bigr]^\top\in\sR^{n\times d}.
\end{align*}
where $c=N/n$ is the compression ratio. Here $N\bmod n = 0$ is assumed, since otherwise the sequence can be padded to.

The above model is an \model instance by letting 
\begin{align}
    \vphi_i=\ve_{\floor{(i-1)/c}+1}/c.
\end{align}

\subsection{Dilated Convolution Attention Patterns}\label{appx:dilated}
The dilated attention pattern is similar to the sliding window attention
and only considers the context within a predefined window.
It differs in that it attends to every other token:
\begin{align}
    \widetilde{\mK}_t=[\vk_{t-2n + 2},\vk_{t-2n + 4}, ...,\vk_{t-2}, \vk_t]^\top.
\end{align}
It can be simulated with two separate queues $\widetilde{\mK}^{\text{odd}}$ and $\widetilde{\mK}^{\text{even}}$:
\begin{align*}
\widetilde{\mK}^{\text{odd}}_t&=
    \begin{cases}
    \mU\widetilde{\mK}_{t-1}^{\text{odd}} + \ve_n\otimes \vk_{t}, \quad &\text{if } t \text{ is odd}\\
    \widetilde{\mK}_{t-1}^{\text{odd}}, \quad & \text{otherwise}
    \end{cases}
    \\ 
\widetilde{\mK}^{\text{even}}_t&=
    \begin{cases}
    \mU\widetilde{\mK}_{t-1}^{\text{even}} + \ve_n\otimes \vk_{t}, \quad &\text{if } t \text{ is even}\\
    \widetilde{\mK}_{t-1}^{\text{even}}, \quad & \text{otherwise}
    \end{cases}
\end{align*}
Likewise for the values.
Depending on $t$, the query attends to one of the two queues:
$\operatorname{output}=$
\begin{align*}
    \begin{cases}
    \bigl(\widetilde{\mV}^{\text{odd}}\bigr)^\top\softmax(\widetilde{\mK}^\text{odd}\vq_t), \quad &\text{if } t \text{ is odd}\\
    \bigl(\widetilde{\mV}^{\text{even}}\bigr)^\top\softmax(\widetilde{\mK}^\text{even}\vq_t), \quad & \text{otherwise}.
    \end{cases}
\end{align*}
The above implementation could incur considerable amount of overhead
and may be actually more expensive than the the original dilated window formulation.
Therefore it has more conceptual value than practical value.

\subsection{Shared Workspace and Linear Unified Nested Attention}\label{appx:luna}
\rev{
Concurrently to this work,
shared workspace~(SW;~\citealp{goyal2021sw}) and linear unified nested attention (LUNA;~\citealp{ma2021luna}) also propposed methods to learn contextualized memory control strategies.
Both can be seen as instances of \model.
At layer $\ell$, their $\vphi_i^{\ell}$ is a function of previous layer's memory $\widetilde{\mX}^{\ell-1}\in\sR^{n\times d}$ and current layer's input $\mX^{\ell}\in\sR^{N\times d}$:
\begin{align}
    \vphi_i = \left[\softmax\left(\widetilde{\mX}^{\ell-1} {\mX^{\ell}}^\top\right)\right]_{:, i},
\end{align}
where $[\boldsymbol{\cdot}]_{:, i}$ denotes the $i$th column of a matrix.
Query, key, and value projections are suppressed for notation clarity.
}

\rev{
SW and LUNA reveal the entire sequence to the control vectors,
by constructing $\vphi$ as a function of previous layer's memory.
Although both admit the recurrent computation as all \model models do,
they are ill-suited for causal attention and autoregressive decoding, 
since future information is ``leaked'' to $\vphi_i$ from the previous layer.
LUNA resorts to a variant of \citet{katharopoulos20transformers} in causal attention~\citep{ma2021luna}.
In contrast, \modelmlp never conditions $\vphi_i$ on previous layer's memory,
but only on the current layer's input.
}

\section{More Details about \model-MLP}\label{appx:abc_mlp}
\subsection{Normalization in Causal Attention}\label{appx:causal_abc}
An equivalent implementation to Eq.~\ref{eq:phi_mlp} is to normalize $\widetilde{\mK}$ and $\widetilde{\mV}$
instead of $\vphi_i$ vectors:
\begin{align*}
    \valpha_i &= \vexp\left(\mW_{\vphi}\vx_i\right),\quad
    \vphi_i = \valpha_i, \\
    \bar{\mK} &= \widetilde{\mK} \left/ \sum_{j=1}^{N}\valpha_j.\right.  
    \quad
    \bar{\mV} = \widetilde{\mV} \left/ \sum_{j=1}^{N}\valpha_j.\right.\\
    \operatorname{output}&=\bar{\mV}^\top\softmax(\bar{\mK}\vq).
\end{align*}
$\mM/\vz$ divides the $\ell$th row of matrix $\mM$ by vector $\vz$'s $\ell$th dimension. 
This admits a linear complexity computation graph for the causal variant of \modelmlp.

\subsection{Higher-Dimensional Case of Example 1}\label{appx:high_dimensional}
This section generalizes Example~\ref{example:exp_phi}
to higher dimensional cases.
Assume that the constant-sized memory has $n$ slots.
$\vphi_i$ is cauculated as in Eq.~\ref{eq:phi_mlp}.
Then $\widetilde{\mK}
    = \sum_{i=1}^{N}\vphi_i\otimes\vk_i\in \sR^{n\times d}$.
Each row of $\widetilde{\mK}$ can be seen
as a separate attention mechanism with a pseudo query.
Let $[\boldsymbol{\cdot}]_\ell$ denote the $\ell$th row/dimension
of a matrix/vector. Then for any $\ell=1,\dots,n$,
\begin{align*}
    \bigl[\widetilde{\mK}\bigr]_\ell
    &=\sum_{i=1}^{N}[\vphi_i]_\ell\otimes\vk_i\\
    &=\sum_{i=1}^{N}\frac{\exp([\mW_\phi]_\ell\cdot\vx_i)}{\sum_{j=1}^{N}\exp([\mW_\phi]_\ell\cdot\vx_j)}\vk_i^\top\\
    &=\attn\left([\mW_\phi]_\ell,\{\vx_i\}_{i=1}^{N}, \{\vk_i\}_{i=1}^{N}\right)^\top\in \sR^{1\times d}.
\end{align*}
In other words, there are $n$ attention mechanisms in total,
each with a separately-parameterized pseudo-query $[\mW_\phi]_\ell$.
They summarize the context for $n$ times in parallel, each producing a $d$-dimensional
vectors.
These output vectors are then stacked into $n$-by-$d$ memory $\widetilde{\mK}$.
$\widetilde{\mV}$ is similar.

\begin{table*}[t]
\centering
\begin{tabular}{@{} l @{\hspace{15pt}}  l ccc c@{}}
\toprule[.1em]
\textbf{Model} 
& $\vphi$
& \textbf{Cross $n$}
& \textbf{Causal $n$}
& \textbf{Encoder $n$}
& \textbf{BLEU}\\

\midrule[.1em]
\base & - & -& - & - & 27.2 \\
\midrule[.1em]
\multirow{10}{*}{\model}
& Window & 32 & 32 & - &26.3 \\
\cmidrule{2-6}
& Cluster & 32 & 32 & - &26.8 \\

\cmidrule{2-6}
& MLP-$\operatorname{ReLU}$ & 32 & \phantom{0}8 & - & -\\
& MLP-$\operatorname{ReLU}$ & 32 & 32 & - & 26.4\\

\cmidrule{2-6}
& MLP-$\operatorname{sigmoid}$ & 32 & \phantom{0}8 & -& 26.8\\
& MLP-$\operatorname{sigmoid}$ & 32 & 32 & - & 27.0\\

\cmidrule{2-6}
& MLP-$\exp$ & 32 & \phantom{0}8 & - & 27.1\\
& MLP-$\exp$ & 32 & 32 & - & \textbf{27.3}\\
 
& MLP-$\exp$-all & 32 & 32 & 32 & 27.0\\

\bottomrule[.1em]
\end{tabular}
\caption{\label{tab:wmt_additional}
\model variants' performance (SacreBLEU) on the WMT14 EN-DE test set for sentence-level machine translation.
MLP-$\operatorname{ReLU}$ with 32/8 memory sizes fails to converge.
 MLP-$\exp$-all applies \model in both the encoder and the decoder,
 while others only in the decoders.
} 
\end{table*}

\section{Experimental Details} \label{appx:experiments}

\subsection{Language Modeling}
We closely build on \citet{baevski2018adaptive}
and \citet{kasai2021t2r}.
The hyperparameters are summarized in Table~\ref{tab:lm_hyperparams}.
All models are trained on 4 A100 GPUs.

\subsection{Machine Translation} \label{appx:mt}
We experiment with a sentence-level (WMT14 EN-DE, \citealp{bojar2014wmt}) and a document-level benchmark (IWSLT14 ES-EN, \citealp{cettolo2014report}) to assess model performance over various sequence lengths.
The preprocessing and data splits of WMT14 EN-DE follow \citet{vaswani2017attention}.
A 32,768 byte pair encoding (BPE;~\citealp{sennrich2016bpe}) vocabulary is shared between source and target languages.
For IWSLT14, we follow \citet{miculicich2018document} and use the \emph{dev2010} subset for development and \emph{tst2010-2012} for testing. The tokenization is also the same as \citet{miculicich2018document}: we tokenize and truecase Spanish and English with Moses~\citep{koehn-etal-2007-moses} and run byte-pair encoding with 30k splits, shared between the two languages. The final dataset contains 1421, 8, and 42 documents for training, development, and testing. On average, each document contains 126.7 sentences, and each sentence contains 21.7(ES)/22.5(EN) BPE subwords. We use a sliding window with length-4 and stride-one to generate our dataset. 
During inference, we use predicted context on the target side.

We average the checkpoints from the last five epochs to obtain the final model \cite{vaswani2017attention}.
In inference, we apply beam search with size 5 and length penalty 0.6.
Other hyperparameters are summarized in Table~\ref{tab:mt_hyperparams}.
All models are trained on 4 RTX 2080 Ti GPUs.

\paragraph{Additional machine translation results.}
In addition to the results presented in \S\ref{sec:mt},
Table~\ref{tab:wmt_additional} further compares,
on the WMT14 EN-DE dataset,
the clustering-based (\S\ref{sec:cluster})
and sliding-window (\S\ref{sec:sw}) models of \model,
as well as  $\relu$ and $\operatorname{sigmoid}$ variants of \modelmlp.
Clustering and sliding-window \model variants \emph{underperform} \modelmlp with the same memory sizes by more than 0.5 BLEU.
Both $\relu$ and $\operatorname{sigmoid}$ \emph{underperform} their $\exp$ counterpart.

\rev{
MLP-$\exp$-all
replaces the encoder's softmax attention modules with \model, in addition to the decoder's.
It underperforms \modelmlp by only 0.3 BLEU.}

\rev{Figure~\ref{fig:memory_length} compares \modelmlp's (32-8 memory sizes) attention memory overhead with softmax attention's.
Following \citet{kasai2021t2r}, we consider a synthetic sequence-to-sequence generation task with varying sequence lengths.
A batch size of 16 and greedy decoding is used.
The models are of the same size as those in \S\ref{sec:mt}.
}

\begin{table*}[th]
\centering
	\centering
	\begin{tabular}{@{}l  rrrrr@{}} 
		\toprule
		
		\textbf{Data} & \textbf{Train} & \textbf{Dev.} & \textbf{Test} & \textbf{Vocab.} & \textbf{Sent./doc}\\ 
		
		\midrule
		WikiText-103 & 103M & 218K & 246K & 268K & -\\
		\midrule
		WMT14 EN-DE & 4.5M & 3K & 3K & \phantom{/K}32K & -\\
		\midrule
		IWSLT14 ES-EN & 1713 & 8& 56 & 30K & 121.5\\
		
		\bottomrule
	\end{tabular}
	\caption{Statistics for the datasets.
	    WikiText-103 split sizes are in number of tokens,
	    WMT14 in number of sentences,
	    and IWSLT14 in number of documents.
	}
	\label{tab:data}
\end{table*}
\subsection{Masked Language Model Finetuning}
Our data for continued pretraining is a
concatenation of
BookCorpus~\citep{zhu2015aligning},
English Wikipedia,
OpenWebText~\citep{gokaslan2019open},
and RealNews~\citep{zellers2019defending}.
Our data differs from RoBERTa's pretraining data, which we do \emph{not} have access to.
We replace their CC-News~\citep{nagel2016news} with RealNews, and drop Stories~\citep{trinh2018simple}.
At the time of this project, the public access to the Stories dataset is broken.\footnote{\url{https://console.cloud.google.com/storage/browser/commonsense-reasoning/reproduce/stories_corpus?pli=1}}
Our machine does \emph{not} have a large enough memory to load
all the data.
We therefore split the training data into 20 shards, after shuffling.
Other preprocessing is the same as \citet{liu2019roberta}.\footnote{
\url{https://github.com/pytorch/fairseq/blob/master/examples/roberta/README.pretraining.md}
}
The hyperparameters for continued pretraining follow base-sized RoBERTa,
part of which are summarized in Table~\ref{tab:mlm_hyperparams}.
All models are trained on a single TPU v3 accelerator.

For downstream task finetuning, we use the same hyperparameters as \citet{liu2019roberta}.\footnote{
\url{https://github.com/pytorch/fairseq/blob/master/examples/roberta/README.glue.md}
}
Table~\ref{tab:glue_data} briefly describes the tasks. The readers
are referred to \citet{wang2018glue} for futher details.

\begin{table}[th]
\centering
\begin{tabular}{@{} l rr@{}}
\toprule[.1em]

\textbf{Hyperprams.} 
& B\&A  %\citet{baevski2018adaptive} 
& Kasai \\ %\citet{kasai2021t2r}\\

\midrule[.1em]
\# Layers & 16 &32\\
\# Heads & 8 & 8\\
Embedding Size  & 1024 & 1024\\
Head Size & 128 & 128\\
FFN Size & 4096 & 4096 \\
Batch Size & 64 & 64\\
Learning Rate & 1.0 & 1.0 \\
Dropout & 0.3 & 0.3\\
Layer Dropout & - & 0.2 \\ 
Memory size & $[32, 64]$ & 64\\
\bottomrule[.1em]
\end{tabular}
\caption{\label{tab:lm_hyperparams}
Hyperparameters used in the language modeling experiments.
B\&A: \citet{baevski2018adaptive};
Kasai: \citet{kasai2021t2r}.
} 
\end{table}

\begin{table}[th]
\centering
\begin{tabular}{@{} l rr@{}}
\toprule[.1em]

\textbf{Hyperprams.} & \textbf{WMT14} & \textbf{IWSLT14}\\

\midrule[.1em]
\# Layers & 6 & 6\\
\# Heads & 8 & 8\\
Embedding Size  & 512 & 512\\
Head Size & 64 & 64\\
FFN Size & 2048 & 1024\\
Warmup Steps & 6000 & 4000\\
Dropout & 0.1 & 0.3\\
Cross Attn. $n$ & 32 & 128\\
Causal Attn. $n$ & 8 & 64\\

\bottomrule[.1em]
\end{tabular}
\caption{\label{tab:mt_hyperparams}
Hyperparameters used in the machine translation experiments.} 
\end{table}

\begin{table}
\centering
\begin{minipage}[b]{.48\textwidth}
\centering
\begin{tabular}{@{} l r@{}}
\toprule[.1em]

\textbf{Hyperprams.} & \textbf{Values} \\
\midrule[.1em]
\# Layers & 12\\
\# Heads & 12\\
Embedding Size  & 768\\
Head Size & 64\\
FFN Size & 3072\\
Dropout & 0.1 \\
Memory Size & $[64, 128]$\\

\bottomrule[.1em]
\end{tabular}
\vspace{8pt}
\caption{\label{tab:mlm_hyperparams}
Hyperparameters for continued pretraining in the masked language model finetuning experiments.} 
\end{minipage}%
\hfill
\begin{minipage}[b]{.48\textwidth}
\centering
\begin{tabular}{@{} l  l  r  r @{}}
\toprule
\textbf{Data} & \textbf{Task} & \textbf{Train} & \textbf{Dev.} \\
\midrule
\textbf{MNLI} & Entailment & 392K & 9.8K \\
\textbf{QNLI} & Entailment & 105K & 5.5K \\
\textbf{QQP} & Paraphrase & 363K & 40K \\
\textbf{SST-2} & Sentiment & 67K & 873 \\
\bottomrule
\end{tabular}
\vspace{8pt}
\caption{\label{tab:glue_data} GLUE datasets and statistics.
MNLI: \citet{williams2017broad};
QNLI is compiled by GLUE's authors using \citet{rajpurkar-etal-2016-squad};
QQP: \citet{qqp};
SST-2: \citet{socher-etal-2013-recursive}.
}
\end{minipage}
\end{table}

\end{appendices}

\end{document}